\documentclass[sigconf]{acmart}

\usepackage{algorithm}
\usepackage{algorithmic}
\usepackage{multirow}
\usepackage{booktabs}
\usepackage{xcolor}
\usepackage{colortbl}
\usepackage{makecell}
\usepackage{diagbox}
\usepackage{amsmath}

\AtBeginDocument{%
  }

\setcopyright{none}
\renewcommand\footnotetextcopyrightpermission[1]{}
\acmYear{2026}
\copyrightyear{2026}
\acmConference[CIKM '26]{ACM International Conference on Information and Knowledge Management}{2026}{Anonymous Review}
\acmBooktitle{ACM International Conference on Information and Knowledge Management (CIKM '26), 2026, Anonymous Review}
\acmDOI{}
\acmISBN{}

\newcommand{\ours}{FedCGR}

\newcommand{\best}[1]{\textbf{#1}}
\newcommand{\second}[1]{\underline{#1}}

\begin{document}

\title{FedCGR: Federated Cross-Domain Generative Recommendation}

\author{Zhuodong Liu}
\affiliation{%
  \institution{Beijing Jiaotong University}
  \country{China}
}
\email{22711104@bjtu.edu.cn}

\author{Hugen Lv}
\affiliation{%
  \institution{Beijing Jiaotong University}
  \country{China}
}
\email{23722056@bjtu.edu.cn}

\author{Xiangyu Li}
\affiliation{%
  \institution{Shanghai Jiao Tong University}
  \country{China}
}
\email{xiangyuli@sjtu.edu.cn}

\author{Bohan Guo}
\affiliation{%
  \institution{University of Malaya}
  \country{Malaysia}
}
\email{23114009@siswa.um.edu.my}

\author{Peiyu Hu}
\authornote{Corresponding author.}
\affiliation{%
  \institution{Xi'an Jiaotong-Liverpool University}
  \country{China}
}
\email{peiyuhu30@gmail.com}

\begin{abstract}
Cross-domain recommendation (CDR) transfers preference knowledge across related domains, but federated deployment makes cross-domain alignment difficult because the behavioral anchors that align item spaces, such as overlapping users and shared interaction signals, are often sparse, unavailable, or privacy-sensitive across clients.
To address this tension, we revisit federated CDR as generation over a stable semantic item language.
By representing items as discrete semantic ID (SID) sequences derived from public item-side metadata, cross-domain item alignment is induced by a shared vocabulary rather than by exchanging private interactions or aligning domain-specific embeddings.
Directly federating SID-based generators, however, introduces two design constraints: the SID tokenizer must remain fixed to preserve cross-client token consistency, which creates a semantic-only bottleneck because local collaborative filtering (CF) signals cannot be globally shared or aligned; meanwhile, standard federated averaging can cause negative transfer under domain heterogeneity.
To overcome these constraints, we propose \ours{}, a federated generative CDR framework that keeps the item language stable and makes adaptation explicit.
\ours{} injects local CF evidence through a reliability-aware semantic interface and trains a prototype-personalized generator that selectively aggregates shared parameters according to domain relatedness while keeping domain-specific quantities local.
Experiments on six Amazon cross-domain scenarios show that \ours{} consistently outperforms federated generative baselines and achieves competitive performance against strong sequential and federated CDR methods under both full-ranking and sampled evaluation protocols.
Our code is publicly available at \url{https://github.com/ZhuodLiu/FedCGR}.
\end{abstract}

\ccsdesc[500]{Information systems~Recommender systems}
\ccsdesc[300]{Information systems~Information retrieval}
\ccsdesc[300]{Computing methodologies~Machine learning}

\keywords{Cross-Domain Recommendation, Federated Learning, Generative Recommendation, Semantic ID, Personalized Federation}

\maketitle

\section{Introduction}

Cross-domain recommendation (CDR) transfers knowledge across related domains to alleviate sparsity and cold-start issues~\cite{zang2022survey}.
Existing CDR methods have made substantial progress through embedding mapping, cross-domain feature interaction, and transferable user representation learning, including CoNet~\cite{conet}, DDTCDR~\cite{ddtcdr}, CATN~\cite{catn}, RecGURU~\cite{recguru}, and DisenCDR~\cite{disencdr}.
However, many of these methods assume centralized access to cross-domain interactions, which is difficult to satisfy when user behavior is held by different platforms or business units.
Federated Learning (FL)~\cite{fedavg} offers a natural alternative by allowing clients to collaboratively train models while keeping raw interaction sequences local.
Yet federated CDR faces a fundamental tension: the behavioral anchors that make cross-domain item spaces alignable---overlapping users, co-occurrence graphs, or shared interaction signals---are precisely the signals that become sparse, unavailable, or privacy-sensitive under federated isolation.

\begin{figure}[t]
  \centering
  \includegraphics[width=\columnwidth]{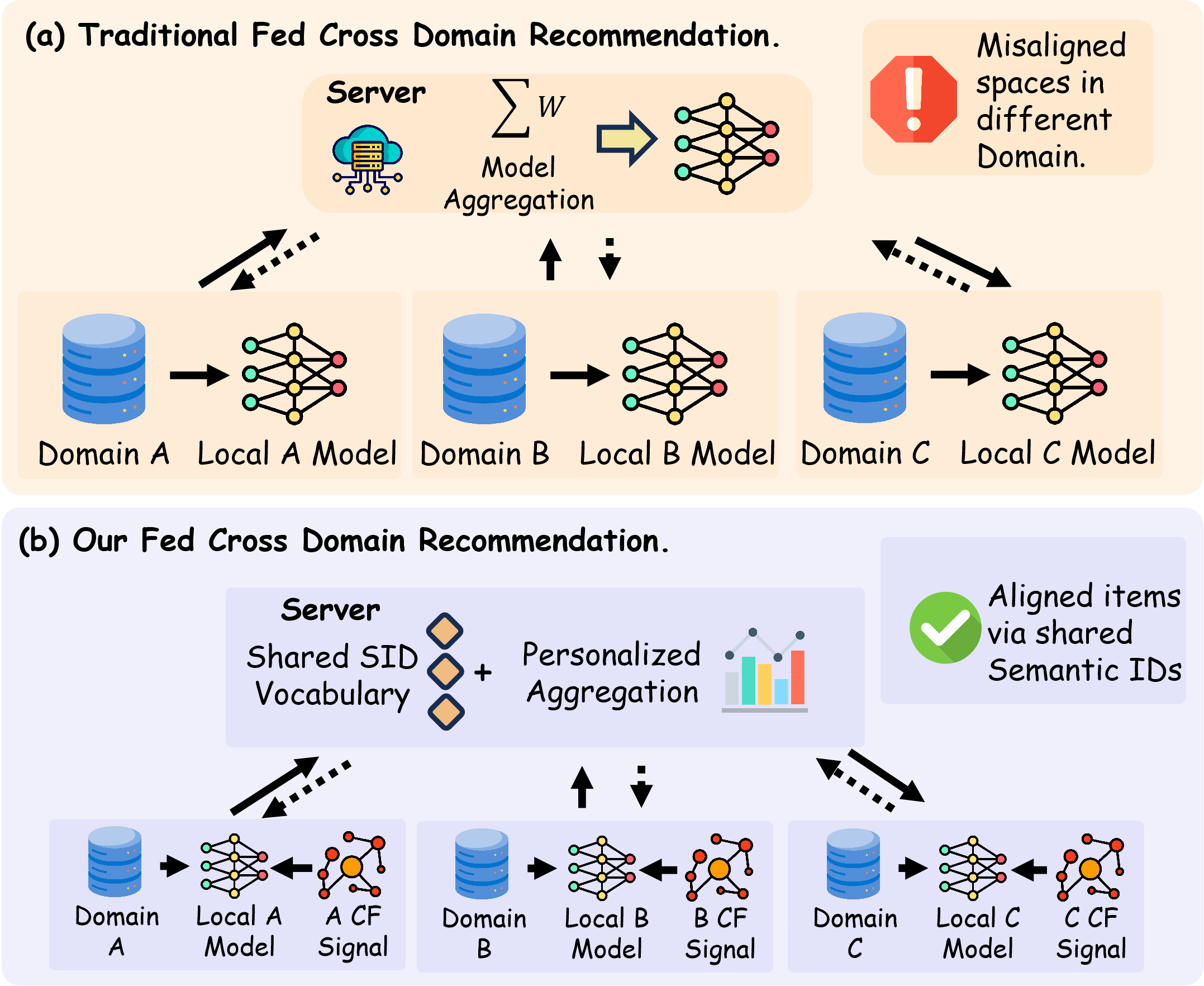}
  \caption{Comparison between (a)~traditional federated CDR, where domain-specific representations are misaligned across clients, and (b)~our approach, which aligns items through a shared SID vocabulary with personalized aggregation, while each client enriches representations with local CF signals that remain private.}
  \label{fig:motivation}
  \vspace{-4mm}
\end{figure}

Federated CDR methods have explored peer-to-peer collaboration~\cite{p2fcdr}, disentangled sequence modeling~\cite{feddcsr}, graph or hypergraph modeling~\cite{fedgcdr,fedhcdr}, and prototype-based contrastive learning~\cite{fedpclcdr}.
Despite these advances, most existing federated CDR models are still discriminative: they score candidate items in continuous representation spaces that are often domain-specific.
As shown in Figure~\ref{fig:motivation}(a), cross-domain transfer therefore depends on making these private spaces comparable, creating a coupling between federated privacy and alignment effectiveness.
As illustrated in Figure~\ref{fig:motivation}(b), SID-based generative recommendation changes what is shared across domains.
Instead of forcing clients to align private continuous item spaces, it represents each item as a discrete token sequence produced by a shared tokenizer trained on public item-side metadata such as title, category, and description~\cite{tiger}.
This effectively defines a shared \emph{item language}: cross-domain item alignment becomes a property of the shared discrete vocabulary rather than an additional objective built on private behavioral data.
Centralized generative CDR has already demonstrated the promise of unified semantic tokenization~\cite{gencdr}, but centralized interaction access is incompatible with the federated setting.

However, a shared SID vocabulary is not a plug-and-play solution for federated CDR.
Directly federating SID-based generators introduces two design constraints.
First, the SID tokenizer must remain fixed to preserve cross-client token consistency; otherwise, the same token may refer to inconsistent semantic regions across domains.
Fixing the tokenizer, however, creates a \emph{semantic-only bottleneck}: domain-specific collaborative filtering (CF) signals cannot be incorporated through the tokenizer itself, and local CF embeddings trained independently on different clients are noisy, non-uniform in quality, and not coordinate-aligned across domains.
Second, standard federated averaging is inadequate for autoregressive SID generation under domain heterogeneity.
Since SID prediction decomposes an item into multiple sequential token decisions, perturbations in early token distributions can cascade through the prefix trie and cause negative transfer.
Therefore, federated generative recommendation requires not only a shared token space, but also reliability-aware behavior injection and similarity-aware parameter sharing.

We propose \ours{}, a federated cross-domain generative recommendation framework organized around one principle: \emph{keep the item language stable and make adaptation explicit}.
The stable part is a fixed SID tokenizer trained from item-side metadata, providing a shared discrete vocabulary across all clients.
The adaptive part is implemented at two levels.
First, \ours{} builds a \emph{reliability-aware semantic interface}: each interacted item is represented by its SID token embedding plus a residual CF signal whose strength is controlled by item-level reliability and a client-local gate.
Second, \ours{} trains a \emph{prototype-personalized generator}: shared generator parameters are aggregated more strongly from behaviorally related domains, while domain embeddings, gates, private experts, and local CF statistics stay on each client.
This design merges what could appear as many components into two mechanisms that each address one of the two constraints above: reliable local evidence on a stable semantic interface, and personalized federated generation over that interface.

The main contributions of this paper are summarized as follows:
\begin{itemize}
  \item We formulate federated CDR as generation over a stable semantic item language, where public item-side metadata defines a shared SID vocabulary while raw interactions remain local.

  \item We propose \ours{}, which addresses two design constraints of federated SID generation---the semantic-only bottleneck from frozen tokenization, and negative transfer from uniform aggregation under domain heterogeneity.

  \item \ours{} introduces a reliability-aware semantic interface that injects local CF evidence through item-level confidence and client-local gating, and a prototype-personalized generator with shared-private experts that selectively transfers parameters according to domain relatedness.

  \item Experiments on six Amazon cross-domain scenarios show that \ours{} consistently improves over federated generative baselines, is competitive with strong sequential recommenders under full-ranking, and achieves the best results among federated CDR methods under sampled evaluation. Ablations reveal a cross-over pattern where CF enrichment and personalized aggregation become the binding constraint under different relatedness regimes.
\end{itemize}


\section{Preliminaries}

\subsection{Federated CDR}
We consider $K$ domains $\{\mathcal{D}_1,\dots,\mathcal{D}_K\}$ in a \emph{cross-silo} federated setting, where each client corresponds to a domain, platform, or business unit rather than an individual user device.
This setting matches practical CDR scenarios in which data silos are organizationally separated and cross-domain interactions cannot be centralized.
Each domain $\mathcal{D}_i$ maintains a user set $\mathcal{U}_i$, an item set $\mathcal{V}_i$, and private interaction sequences $\mathcal{R}_i$.
Raw user-item interactions and user sequences are not shared across domains.
The goal is to train a domain-specific next-item recommender $p_{\theta_i}(v_{T+1}\mid S_u)$ for each domain while leveraging transferable knowledge from other domains through federated parameter sharing.
In standard FedAvg~\cite{fedavg}, shared parameters are aggregated by data volume:
\begin{equation}
  W_{\mathrm{sh}}^{(t+1)}=\sum_{i=1}^{K}\frac{n_i}{\sum_j n_j}W_{i,\mathrm{sh}}^{(t)},
\end{equation}
where $n_i$ is the number of local training instances.
\ours{} replaces this single global aggregate with a personalized aggregate for each domain.

\subsection{SID-Based Generative Recommendation}
Given a user sequence $S_u=(v_1,\dots,v_T)$, SID-based generative recommendation~\cite{tiger,survey-genrec} represents each item $v$ with a discrete code sequence $\mathbf{c}_v=(c_v^1,\dots,c_v^L)$.
The codes are produced by Residual Quantization (RQ)~\cite{rqvae} over item-side features.
At level $l$, the tokenizer selects the closest codeword:
\begin{equation}
  c_v^l=\arg\min_j\|\mathbf{r}_v^l-\mathbf{z}_j^l\|_2,\quad
  \mathbf{r}_v^{l+1}=\mathbf{r}_v^l-\mathbf{z}_{c_v^l}^l.
\end{equation}
The recommender predicts the next item by autoregressively generating its SID:
\begin{equation}
  p(v_{T+1}\mid S_u)=\prod_{l=1}^{L}p_\theta(c_{T+1}^{l}\mid S_u,c_{T+1}^{1:l-1}).
\end{equation}
During inference, generated SIDs are mapped back to valid items in the target domain through a prefix trie.

\section{Methodology}

\subsection{Overview}
\label{sec:overview}
\begin{figure*}[t]
  \centering
  \includegraphics[width=0.95\textwidth]{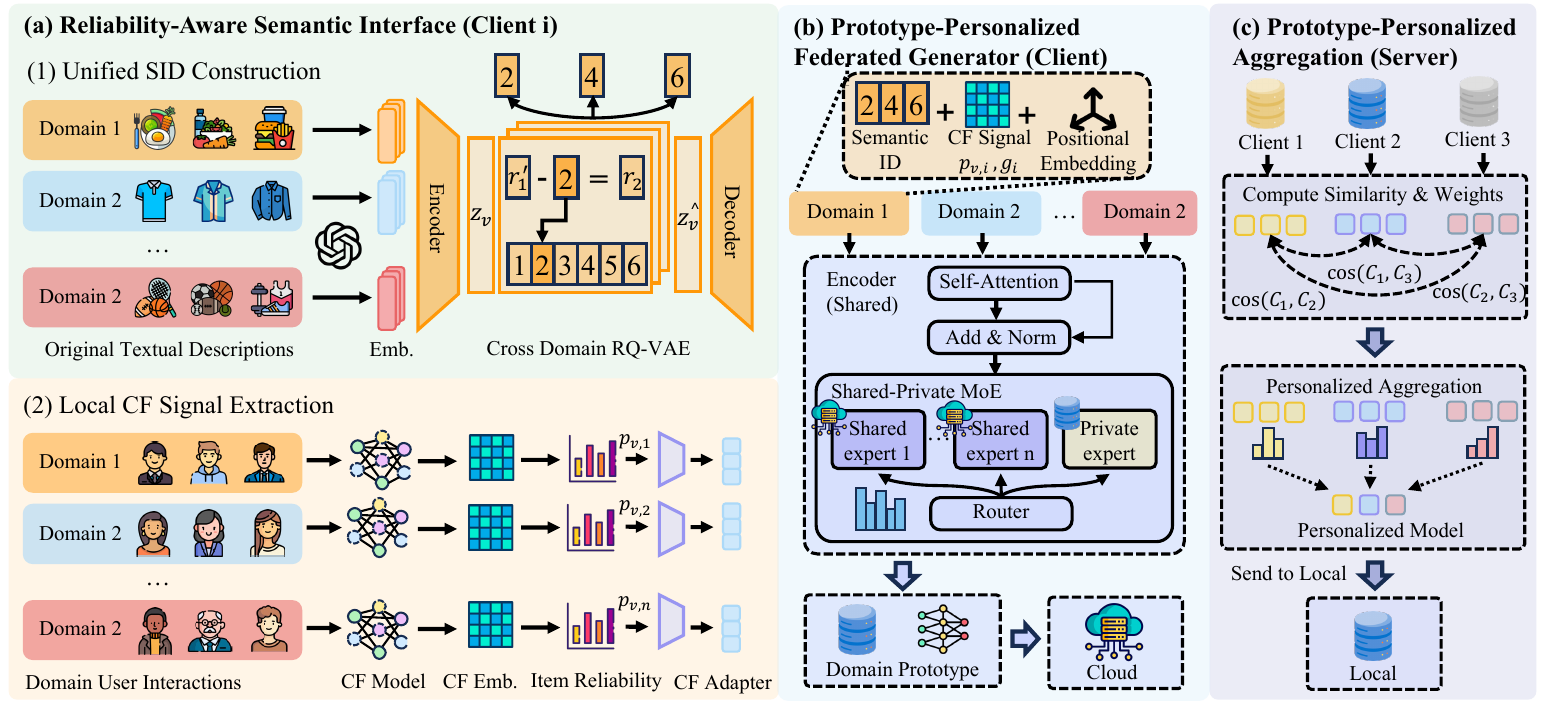}
  \caption{The overall framework of \ours{}.
  A fixed RQ-VAE tokenizer maps item-side metadata into shared SIDs, forming a stable vocabulary across domains.
  Each client locally extracts CF evidence and reliability scores from private interactions, and injects reliable CF information into SID representations through a client-local behavior adapter.
  The autoregressive generator is trained federatively with prototype-personalized aggregation: shared parameters are selectively transferred according to domain relatedness, while domain-specific quantities remain local.}
  \Description{The framework consists of three parts.
  Part~(a), Reliability-Aware Semantic Interface, shows how each client constructs a unified SID vocabulary from item textual descriptions via a cross-domain RQ-VAE encoder, and separately extracts local CF embeddings, item reliability scores, and CF adapter outputs from domain-specific user interactions.
  Part~(b), Prototype-Personalized Federated Generator, depicts the client-side model architecture: semantic ID tokens, CF signals modulated by a gate, positional embeddings, and domain tokens are fed into a shared encoder with self-attention, followed by a shared-private Mixture-of-Experts block containing shared experts and one private expert governed by a router, producing a domain prototype.
  Part~(c), Prototype-Personalized Aggregation on the server, illustrates how the server computes pairwise cosine similarities among client prototypes to derive personalized aggregation weights, constructs a personalized model for each client, and sends the personalized parameters back to each client.}
  \label{fig:framework}
  \vspace{-4mm}
\end{figure*}

Figure~\ref{fig:framework} illustrates \ours{}.
The central idea is to use a fixed SID space as the communication interface among domains, and to make everything built on top of this interface adaptive to reliability and domain relatedness.
Item-side metadata is mapped to SIDs, which provide a shared discrete vocabulary across clients without using private interactions.
Because this vocabulary must remain stable during federated training, \ours{} does not update the SID tokenizer after item-side pre-training.
Instead, each client enriches SID token representations with local CF evidence extracted from its own interaction sequences.
Since CF embeddings are trained independently on different clients, their coordinate systems are not comparable; therefore \ours{} keeps each client's CF adapter and dense auxiliary head local rather than aggregating them.

The residual CF injection is reliability-aware: item frequency estimates whether the local CF embedding is trustworthy, and a client-local gate calibrates the domain-level CF contribution.
The resulting sequence representation is consumed by an autoregressive SID generator whose shared parameters are aggregated more strongly from behaviorally related domains via domain prototypes, while domain-specific quantities remain on each client.
Thus, the method has two coupled parts rather than many independent modules: a \emph{reliability-aware semantic interface} for representing user histories, and a \emph{prototype-personalized generator} for federated transfer.
We note that the frozen RQ-VAE tokenizer only determines item-to-SID assignments; the SID token embedding tables, Transformer parameters, prediction heads, and expert parameters are still trainable and can be optimized and aggregated without changing the SID assigned to any item.

\textbf{Privacy scope.}
\ours{} follows the standard cross-silo FL setting in which raw user sequences and local item-level CF statistics are not shared.
Only shared generator parameters and low-dimensional domain prototypes are uploaded; domain embeddings, CF gates, CF adapters, dense heads, private experts, CF embeddings, and confidence scores remain on each client (Table~\ref{tab:param_partition}).
We do not claim formal differential privacy; secure aggregation and differential privacy are orthogonal mechanisms that can be integrated with \ours{}.

\subsection{Reliability-Aware Semantic Interface}
\label{sec:semantic_interface}

\textbf{Stable SID vocabulary.}
For each item $v$, a frozen pre-trained text encoder, such as BGE (BAAI General Embedding)~\cite{bge}, maps public item-side metadata into $\mathbf{e}_v\in\mathbb{R}^{d_e}$.
An RQ-VAE tokenizer quantizes $\mathbf{e}_v$ into $\mathbf{c}_v=(c_v^1,\dots,c_v^L)$ and is trained with reconstruction, commitment, code utilization, and collision regularization:
\begin{equation}
  \mathcal{L}_{\text{rqvae}}=
  \mathcal{L}_{\text{recon}}+\beta\mathcal{L}_{\text{commit}}
  +\lambda_e\mathcal{L}_{\text{entropy}}+\lambda_c\mathcal{L}_{\text{collision}}.
\end{equation}
After training, the tokenizer is fixed.
This preserves cross-client token consistency: the same SID token always refers to the same semantic region throughout federated training.
Fixing the tokenizer does not freeze the downstream SID token embedding tables used by the recommender.

In implementation, the target-domain prefix trie stores valid item SIDs.
If multiple items in the same domain share a SID, the corresponding trie leaf keeps all associated items.
For a generated SID $\mathbf{c}$, its sequence-level score is computed as
$s_{\mathrm{sid}}(\mathbf{c})=\sum_{l=1}^{L}\log p(c^{l}\mid S_u,c^{1:l-1})$.
All items sharing that leaf receive the same SID score; ties are broken by validation-set item popularity within the target domain.
The collision regularization term $\mathcal{L}_{\text{collision}}$ in the RQ-VAE objective encourages unique SID assignments, so collisions are rare in practice.

\textbf{Local CF evidence.}
The fixed SID tokenizer captures item-side semantics but does not encode domain-specific collaborative co-occurrence.
Therefore, each client $\mathcal{D}_i$ trains a lightweight sequential model (SASRec~\cite{sasrec}; alternatives such as BERT4Rec~\cite{bert4rec} are also applicable) locally using only its private interaction sequences $\mathcal{R}_i$, and extracts a local CF item embedding table:
\begin{equation}
  \{\mathbf{e}^{\mathrm{cf}}_{v,i}\in\mathbb{R}^{d_{\mathrm{cf}}}
  \mid v\in\mathcal{V}_i\}.
\end{equation}
The CF embeddings are kept local and fixed during federated generator training.
Since CF embeddings trained independently on different clients are not guaranteed to be coordinate-aligned, \ours{} does not aggregate the CF adapter across clients.
Instead, each client learns a local behavior adapter $A_i(\cdot)$, analogous to parameter-efficient adaptation modules~\cite{lora}, that maps its own CF embedding space into the hidden space of the SID generator:
\begin{equation}
  A_i(\mathbf{e}^{\mathrm{cf}}_{v,i})
  =
  \operatorname{Dropout}
  \left(
  \operatorname{LayerNorm}
  \left(
  W^{\mathrm{cf}}_i\mathbf{e}^{\mathrm{cf}}_{v,i}
  +\mathbf{b}^{\mathrm{cf}}_i
  \right)
  \right).
  \label{eq:cf_adapter}
\end{equation}
The adapter parameters are client-local and are never uploaded.

Since CF quality varies across items, we define an item-level reliability score from local interaction frequency:
\begin{equation}
  \rho_{v,i}=
  \frac{\log(1+n_{v,i})}
  {\epsilon+\max_{v'\in\mathcal{V}_i}\log(1+n_{v',i})},
  \label{eq:cf_confidence}
\end{equation}
where $n_{v,i}$ is the interaction count of item $v$ in domain $\mathcal{D}_i$, and $\epsilon$ is a small constant for numerical stability.
Head items with sufficient interactions obtain larger $\rho_{v,i}$, while long-tail items receive smaller confidence.

\textbf{Reliability-aware residual fusion.}
For a history item $v_t$ in domain $\mathcal{D}_i$, \ours{} constructs the input representation as
\begin{equation}
  \mathbf{h}_t =
  \sum_{l=1}^{L}\mathbf{E}_{\mathrm{sid}}^l(c_{v_t}^{l})
  +\mathbf{E}_{\mathrm{pos}}(t)
  +\rho_{v_t,i}\cdot g_i\cdot A_i(\mathbf{e}^{\mathrm{cf}}_{v_t,i}),
  \label{eq:fusion}
\end{equation}
where $\mathbf{E}_{\mathrm{sid}}^l$ is the trainable SID token embedding table at level $l$, $\mathbf{E}_{\mathrm{pos}}$ is the positional embedding, and $g_i=\sigma(\alpha_i)$ is a client-local gate.
The residual form allows the model to enrich the stable SID representation with reliable local CF evidence, while suppressing noisy long-tail CF signals through $\rho_{v_t,i}$ and calibrating the domain-level CF contribution through $g_i$.
A domain embedding $\mathbf{e}^{\mathrm{dom}}_i$ is prepended to the sequence:
\begin{equation}
  \mathbf{X}_u=[\mathbf{e}^{\mathrm{dom}}_i;\mathbf{h}_1;\dots;\mathbf{h}_T],
\end{equation}
and remains client-local.

\textbf{Training-only dense regularization.}
To encourage the shared encoder memory to preserve reliable local collaborative evidence, each client uses a local dense prediction head $f_{\mathrm{head},i}(\cdot)$ during training.
Given encoder memories $\mathbf{M}_{1:T}$ over item positions, the head predicts the target item's local CF embedding:
\begin{equation}
  \hat{\mathbf{z}}_n =
  \operatorname{norm}
  \left(
  f_{\mathrm{head},i}
  \left(
  \operatorname{MeanPool}(\mathbf{M}_{n,1:T})
  \right)
  \right),
\end{equation}
where $\operatorname{norm}(\cdot)$ denotes $\ell_2$ normalization.
The target CF embedding is also normalized:
\begin{equation}
  \mathbf{z}_n^+ =
  \operatorname{norm}(\mathbf{e}^{\mathrm{cf}}_{v_n^+,i}).
\end{equation}
The auxiliary loss is a confidence-weighted local InfoNCE (Information Noise-Contrastive Estimation) objective:
\begin{equation}
  \mathcal{L}_{\mathrm{dense}}
  =
  -\frac{1}{|\mathcal{B}|}
  \sum_{n\in\mathcal{B}}
  \rho_{v_n^+,i}
  \log
  \frac{
  \exp(\hat{\mathbf{z}}_n^\top \operatorname{sg}(\mathbf{z}_n^+)/\tau)
  }{
  \sum_{m\in\mathcal{B}}
  \exp(\hat{\mathbf{z}}_n^\top \operatorname{sg}(\mathbf{z}_m^+)/\tau)
  },
  \label{eq:dense_loss}
\end{equation}
where $\operatorname{sg}(\cdot)$ denotes stop-gradient and $\tau$ is the InfoNCE temperature.
Positives and in-batch negatives are drawn only from the local client batch.
The dense head is client-local and is discarded during inference.

\subsection{Prototype-Personalized Federated Generator}
\label{sec:personalized_generator}
The generator is trained federatively, but shared and local quantities are explicitly separated.
The shared parameter set $W_{\mathrm{sh}}$ includes SID token embeddings, encoder/decoder attention layers, prediction heads, shared experts, and shared routing parameters.
Domain embeddings, CF gates, CF adapters, dense heads, local CF statistics, and private experts remain local.

\textbf{Shared-private expert block.}
Each encoder layer uses a shared-private feed-forward block based on a Mixture-of-Experts (MoE) architecture:
\begin{equation}
  \mathbf{H}'=\operatorname{LayerNorm}(\mathbf{H}+\operatorname{MultiHeadAttn}(\mathbf{H})),
\end{equation}
\begin{equation}
  \mathbf{H}''=\operatorname{LayerNorm}(\mathbf{H}'+\operatorname{MoE\text{-}FFN}_i(\mathbf{H}')).
\end{equation}
The block contains $N_s$ shared experts $\{E_1^{\mathrm{sh}},\dots,E_{N_s}^{\mathrm{sh}}\}$ and one client-local private expert $E_i^{\mathrm{pri}}$.
For an input token representation $\mathbf{x}$, the router computes top-$k$ routing weights over the union of shared and private experts:
\begin{equation}
  \operatorname{MoE}_i(\mathbf{x})
  =
  \sum_{e\in\mathcal{T}_k(\mathbf{x})}
  a_e(\mathbf{x})E_e(\mathbf{x}),
  \label{eq:moe}
\end{equation}
where $\mathcal{T}_k(\mathbf{x})$ denotes the selected expert set and $a_e(\mathbf{x})$ is the normalized routing probability.
Shared experts and shared routing parameters participate in personalized aggregation, while the private expert and private routing parameters remain local.
A load-balancing loss encourages stable expert utilization:
\begin{equation}
  \mathcal{L}_{\mathrm{aux}}=N_e\sum_{j=1}^{N_e}f_jp_j,
\end{equation}
where $N_e=N_s+1$, $f_j$ is the fraction of routed tokens, and $p_j$ is the average routing probability for expert $j$ on the local client.

\textbf{Domain prototype.}
At the end of communication round $t$, client $\mathcal{D}_i$ computes a domain prototype from the encoder memories of a local mini-batch subset $\mathcal{S}_i^{(t)}$.
To prevent the prototype from being dominated by the prepended domain token, we average only over item positions:
\begin{equation}
  \bar{\mathbf{p}}_i^{(t)}
  =
  \frac{1}{|\mathcal{S}_i^{(t)}|}
  \sum_{(u,S_u)\in\mathcal{S}_i^{(t)}}
  \operatorname{MeanPool}_{1:T}
  \left(
  \operatorname{sg}(\mathbf{M}_{u,1:T})
  \right).
\end{equation}
The prototype is updated with an exponential moving average (EMA) and normalized:
\begin{equation}
  \mathbf{p}_i^{(t)}
  =
  \operatorname{norm}
  \left(
  \beta\mathbf{p}_i^{(t-1)}
  +(1-\beta)\bar{\mathbf{p}}_i^{(t)}
  \right).
\end{equation}
The prototype is a compact domain-level summary used only for aggregation weighting.

\textbf{Prototype-personalized aggregation.}
After receiving shared parameters $\{W_{j,\mathrm{sh}}^{(t+1)}\}_{j=1}^{K}$ and prototypes $\{\mathbf{p}_j^{(t+1)}\}_{j=1}^{K}$ from clients, the server constructs a personalized aggregate for each target domain:
\begin{equation}
  \bar{W}_{i,\mathrm{sh}}^{(t+1)}
  =
  \sum_{j=1}^{K}
  \omega_{ij}^{(t+1)}
  W_{j,\mathrm{sh}}^{(t+1)},
  \label{eq:pers_agg}
\end{equation}
where
\begin{equation}
  \omega_{ij}^{(t+1)}
  =
  \frac{
  n_j\exp(\cos(\mathbf{p}_i^{(t+1)},\mathbf{p}_j^{(t+1)})/\tau_a)
  }{
  \sum_{k=1}^{K}
  n_k\exp(\cos(\mathbf{p}_i^{(t+1)},\mathbf{p}_k^{(t+1)})/\tau_a)
  }.
  \label{eq:agg_weight}
\end{equation}
Here $n_j$ is the number of local training instances on client $\mathcal{D}_j$, and $\tau_a$ controls the sharpness of similarity-aware aggregation.
When $\tau_a\rightarrow\infty$, the similarity term becomes uniform and the aggregation reduces to data-volume-weighted FedAvg.

\subsection{Autoregressive SID Prediction and Training}
\label{sec:decoder}
At SID level $l$, the decoder input is
\begin{equation}
  \mathbf{q}^l=\begin{cases}
    \mathbf{e}_{\mathrm{sos}}, & l=1,\\
    f_{\mathrm{ctx}}^l([\mathbf{E}_{\mathrm{sid}}^1(c^1);\dots;\mathbf{E}_{\mathrm{sid}}^{l-1}(c^{l-1})]), & l>1,
  \end{cases}
\end{equation}
where $\mathbf{e}_{\mathrm{sos}}$ is a learnable start token.
After adding a level embedding $\mathbf{e}_{\mathrm{lvl}}^l$, the Transformer decoder attends to the encoder memory $\mathbf{M}$ and predicts the next token:
\begin{equation}
  p(c^l\mid S_u,c^{1:l-1})=
  \operatorname{softmax}(W_{\mathrm{head}}^l\hat{\mathbf{q}}^l+\mathbf{b}_{\mathrm{head}}^l).
\end{equation}
The SID generation loss is
\begin{equation}
  \mathcal{L}_{\mathrm{sid}}=
  -\frac{1}{L}\sum_{l=1}^{L}
  \log p_{\theta}(c_{T+1}^{l}\mid S_u,c_{T+1}^{1:l-1}).
\end{equation}
For client $\mathcal{D}_i$, the local training objective is
\begin{equation}
  \mathcal{L}_i
  =
  \mathcal{L}_{\mathrm{sid}}
  +\lambda_d\mathcal{L}_{\mathrm{dense}}
  +\lambda_a\mathcal{L}_{\mathrm{aux}}
  +\frac{\mu}{2}
  \left\|
  W_{i,\mathrm{sh}}-\bar{W}_{i,\mathrm{sh}}^{(t)}
  \right\|_2^2.
  \label{eq:total_loss}
\end{equation}
The proximal term is applied only to shared generator parameters.
Client-local quantities, including domain embeddings, CF gates, CF adapters, dense heads, private experts, private routing parameters, CF embeddings, and confidence scores, are optimized or maintained locally and are not uploaded.

\begin{table}[t]
  \caption{Parameter and statistic partition in \ours{}.
  The SID tokenizer is fixed after item-side pre-training; only shared generator parameters participate in federated aggregation.}
  \label{tab:param_partition}
  \centering
  \small
  \setlength{\tabcolsep}{3pt}
  \begin{tabular}{@{}llc@{}}
    \toprule
    \textbf{Category} & \textbf{Parameters / Quantities} & \textbf{Comm.} \\
    \midrule
    Frozen tok.
    & RQ-VAE encoder/codebooks, item-to-SID map
    & Not updated \\
    \midrule
    \multirow{5}{*}{\makecell[l]{Federated\\shared}}
    & Encoder/decoder attention    & Pers.\ agg. \\
    & SID token embeddings         & Pers.\ agg. \\
    & Prediction heads             & Pers.\ agg. \\
    & Shared experts               & Pers.\ agg. \\
    & Shared router params         & Pers.\ agg. \\
    \midrule
    \multirow{6}{*}{Client-local}
    & Domain emb.\ $\mathbf{e}^{\mathrm{dom}}_i$
    & Not uploaded \\
    & CF gate $g_i$ / logit $\alpha_i$
    & Not uploaded \\
    & CF adapter $A_i$
    & Not uploaded \\
    & Dense head $f_{\mathrm{head},i}$
    & Train-only \\
    & Private expert
    & Not uploaded \\
    & Private router params
    & Not uploaded \\
    \midrule
    \multirow{2}{*}{\makecell[l]{Local\\stats}}
    & CF embeddings $\mathbf{e}^{\mathrm{cf}}_{v,i}$
    & Kept local \\
    & Confidence scores $\rho_{v,i}$
    & Kept local \\
    \bottomrule
  \end{tabular}
\end{table}

\begin{algorithm}[t]
\caption{\ours{} Federated Training}
\label{alg:fedcgr}
\begin{algorithmic}[1]
\REQUIRE Domains $\{\mathcal{D}_1,\dots,\mathcal{D}_K\}$, item metadata, federated rounds $T_{\mathrm{fed}}$, local epochs $E$
\ENSURE Personalized generative recommender for each domain
\STATE Train and fix the RQ-VAE tokenizer from item-side metadata; assign SIDs to all items
\FOR{each domain $\mathcal{D}_i$ in parallel}
    \STATE Train a local SASRec model and extract $\{\mathbf{e}^{\mathrm{cf}}_{v,i}\}$
    \STATE Compute local confidence scores $\{\rho_{v,i}\}$
\ENDFOR
\STATE Initialize shared generator parameters $W_{\mathrm{sh}}^{(0)}$
\STATE Initialize client-local parameters
\STATE Set $\bar{W}_{i,\mathrm{sh}}^{(0)}\leftarrow W_{\mathrm{sh}}^{(0)}$ for all clients
\FOR{$t=0,1,\dots,T_{\mathrm{fed}}-1$}
    \STATE Server sends $\bar{W}_{i,\mathrm{sh}}^{(t)}$ to each client $\mathcal{D}_i$
    \FOR{each domain $\mathcal{D}_i$ in parallel}
        \STATE Initialize local shared parameters from $\bar{W}_{i,\mathrm{sh}}^{(t)}$
        \FOR{local epoch $=1,\dots,E$}
            \FOR{each local batch}
                \STATE Construct reliability-aware inputs by Eq.~\ref{eq:fusion}
                \STATE Encode user history and autoregressively decode the target SID
                \STATE Update shared and local parameters using Eq.~\ref{eq:total_loss}
            \ENDFOR
        \ENDFOR
        \STATE Compute prototype $\mathbf{p}_i^{(t+1)}$
        \STATE Upload shared parameters $W_{i,\mathrm{sh}}^{(t+1)}$ and prototype $\mathbf{p}_i^{(t+1)}$
    \ENDFOR
    \STATE Server computes $\omega_{ij}^{(t+1)}$ by Eq.~\ref{eq:agg_weight}
    \STATE Server forms $\bar{W}_{i,\mathrm{sh}}^{(t+1)}$ by Eq.~\ref{eq:pers_agg}
\ENDFOR
\end{algorithmic}
\end{algorithm}

Only shared generator parameters and domain prototypes are uploaded.
Raw user sequences, local CF embeddings, confidence scores, gates, domain embeddings, CF adapters, dense heads, private experts, and private routing parameters remain local.
At inference time, \ours{} builds a prefix trie over valid item SIDs in the target domain and applies domain-constrained beam search.
The dense auxiliary head is not used during inference.
\section{Experiments}
In this section, we conduct extensive experiments to evaluate \ours{}.
The experiments are designed to answer the following Research Questions (RQs):
\begin{itemize}
  \item \textbf{(Effectiveness) RQ1:} How does \ours{} perform against single-domain, federated discriminative CDR, and federated generative baselines?
  \item \textbf{(Ablation) RQ2:} What is the contribution of each key component, including reliability-aware fusion and prototype-personalized aggregation?
  \item \textbf{(Cold-Start) RQ3:} Does \ours{} improve recommendation for users with sparse interaction histories?
  \item \textbf{(Sensitivity) RQ4:} How sensitive is \ours{} to key hyper-parameters, and what do the learned reliability signals reveal?
  \item \textbf{(Efficiency) RQ5:} What is the communication and training overhead of \ours{}?
\end{itemize}

\subsection{Experimental Setup}

\subsubsection{Datasets.}
\label{sec:datasets}
We evaluate \ours{} on cross-domain scenarios constructed from Amazon Review datasets~\cite{amazonreview}.
We select five product domains---Grocery, Beauty, Sports, Food, and Kitchen---and compose six scenarios that span different domain counts and relatedness levels.

\textbf{Two-domain scenarios.}
FK (Food--Kitchen) pairs two household consumable domains with strong semantic overlap, representing high-affinity transfer;
GB (Grocery--Beauty) combines two moderately related categories, representing medium-affinity transfer;
GS (Grocery--Sports) combines two largely unrelated categories, representing low-affinity transfer.
These three endpoints allow us to measure how domain relatedness affects federated generative recommendation.

\textbf{Multi-domain scenarios.}
FKB (Food--Kitchen--Beauty) mixes three domains where two are closely related (Food and Kitchen) while the third (Beauty) is more distant, testing whether prototype-personalized aggregation can selectively transfer among unevenly related clients.
GBS (Grocery--Beauty--Sports) mixes three domains of varying relatedness within a single federation, further testing selective transfer.
GKBS (Grocery--Kitchen--Beauty--Sports) increases heterogeneity to four domains and tests scalability.
For each domain, we apply 5-core filtering and split each user sequence by leave-one-out evaluation.
Table~\ref{tab:dataset} summarizes the statistics.

\begin{table}[t]
  \caption{Domain statistics after 5-core filtering (upper) and cross-domain scenario composition (lower).
  Overlap denotes the ratio of users appearing in \emph{all} domains of a scenario to the total unique users across those domains, computed after independent per-domain 5-core filtering.}
  \label{tab:dataset}
  \centering
  \small
  \begin{tabular*}{\columnwidth}{@{\extracolsep{\fill}}lrrrrc@{}}
    \toprule
    \textbf{Domain} & \textbf{\#Users} & \textbf{\#Items} & \textbf{\#Inter.} & \textbf{Avg Len} & \textbf{Density} \\
    \midrule
    Food    & 9,215 & 3,742 & 98,866 & 10.73 & 0.29\% \\
    Kitchen & 7,536 & 4,116 & 78,589 & 10.43 & 0.25\% \\
    Grocery & 6,423 & 2,911 & 63,812 & 9.93  & 0.34\% \\
    Beauty  & 8,159 & 4,727 & 82,438 & 10.10 & 0.21\% \\
    Sports  & 7,102 & 5,166 & 68,747 & 9.68  & 0.19\% \\
    \midrule
    \textbf{Scenario} & \multicolumn{4}{l}{\textbf{Domains}} & \textbf{Overlap} \\
    \midrule
    FK   & \multicolumn{4}{l}{Food, Kitchen}                   & 4.2\% \\
    GB   & \multicolumn{4}{l}{Grocery, Beauty}                  & 2.1\% \\
    GS   & \multicolumn{4}{l}{Grocery, Sports}                  & 0.8\% \\
    FKB  & \multicolumn{4}{l}{Food, Kitchen, Beauty}            & 1.5\% \\
    GBS  & \multicolumn{4}{l}{Grocery, Beauty, Sports}          & 0.6\% \\
    GKBS & \multicolumn{4}{l}{Grocery, Kitchen, Beauty, Sports} & 0.3\% \\
    \bottomrule
  \end{tabular*}
\end{table}

\subsubsection{Baselines.}
Because full-ranking and sampled evaluation require different candidate-generation interfaces, and because several federated CDR repositories report results only under the 999-negative protocol, we organize baselines by evaluation protocol.

\textbf{Full-ranking baselines} (Table~\ref{tab:main_fullrank}).
\textbf{(1)~Single-domain sequential recommenders} (GRU4Rec~\cite{gru4rec}, SASRec~\cite{sasrec}) train each domain independently without cross-domain transfer, serving as a lower reference for the value of federation.
\textbf{(2)~Single-domain generative recommender} (TIGER~\cite{tiger}) uses SID-based autoregressive generation but trains per domain, isolating the benefit of semantic tokenization from federation.
\textbf{(3)~Federated generative baselines} (TIGER+FedAvg, TIGER+FedProx) directly federate TIGER with standard aggregation strategies, testing whether na\"ive federation already helps generative models.

\textbf{999-neg baselines} (Table~\ref{tab:main_999neg}).
\textbf{(4)~Local sequential recommenders} (LocalSASRec, LocalCL4SRec, LocalDuoRec) train contrastive or dual-objective sequential models independently per domain.
\textbf{(5)~Federated discriminative CDR methods} (FedSASRec, FedCL4SRec, FedDCSR~\cite{feddcsr}) federate discriminative sequential recommenders with varying degrees of disentanglement, following their original sampled evaluation protocol for direct comparison.

\subsubsection{Evaluation Protocol.}
For each user sequence, the last item is used for testing, the second-to-last item for validation, and the remaining interactions for training.
We report Hit Rate at rank 10 (H@10) and Normalized Discounted Cumulative Gain at rank 10 (NDCG@10, abbreviated N@10).
Our primary evaluation is \textbf{all-user full-ranking}, where each test user is evaluated against the entire item catalogue of the target domain.
For compatibility with prior federated CDR repositories~\cite{feddcsr,fedhcdr}, we additionally conduct \textbf{sampled evaluation with 999 negatives}: each test user ranks the ground-truth item among 999 randomly sampled negatives.
The negative set is sampled once per user and shared across all methods to ensure fair comparison; we use random seed 42 and report results averaged over three independent runs.

\subsubsection{Implementation Details.}
The SID tokenizer uses BGE-base-en-v1.5 as the text encoder ($d_e=768$), $L=3$ quantization levels, and codebook size $C=256$.
The local CF extractor is SASRec with $d_{\text{cf}}=128$, 2 layers, 2 heads, and 15 training epochs per domain.
After local pre-training, the CF item embeddings are fixed and kept on the corresponding client.
The federated generator uses $d_{\text{model}}=256$, 4 attention heads, $N_{\text{enc}}=N_{\text{dec}}=3$, dropout 0.15, and maximum sequence length 50.
The shared-private expert block uses $N_s=2$ shared experts and one private expert per client, with expert hidden size 512 and top-$k=2$ routing.
Shared experts and shared routing parameters participate in personalized aggregation, while private experts and private routing parameters remain local.
Each client uses a local CF adapter $A_i$ implemented as Linear($128\rightarrow256$)+LayerNorm+Dropout(0.1), and a local training-only dense head implemented as Linear($256\rightarrow64$)+GELU+Linear($64\rightarrow128$).
The CF adapter and dense head are optimized locally during federated generator training but are not uploaded to the server.
We set $\lambda_d=0.1$, $\lambda_a=0.01$, InfoNCE temperature $\tau=0.07$, proximal coefficient $\mu=0.01$, prototype momentum $\beta=0.9$, local epochs $E=2$, federated rounds $T_{\text{fed}}=40$, batch size 1024, learning rate $3\times10^{-4}$ with cosine decay and 5-round warmup, label smoothing 0.1, gradient clip norm 1.0, and beam size $B=50$. The aggregation temperature $\tau_a$ is tuned on validation data. All experiments are repeated with three random seeds (42, 123, 456); standard deviations are below 0.003 on all main metrics and are omitted for readability. All experiments are conducted on NVIDIA RTX 4090 GPUs.

\subsection{Overall Performance (RQ1)}
\label{sec:main_results}

We evaluate \ours{} from two complementary perspectives.
Table~\ref{tab:main_fullrank} reports per-domain results under \emph{full-ranking evaluation}, where each test user is evaluated against the entire item catalogue of the target domain.
Table~\ref{tab:main_999neg} adopts the \emph{999-neg sampled protocol} used by prior federated CDR work~\cite{feddcsr,fedhcdr}, enabling direct comparison with federated discriminative baselines under their own evaluation paradigm.

\begin{table*}[t]
  \centering
  \caption{Per-domain results under full-ranking evaluation (H@10 and N@10).
  Best result per row is \textbf{bold}; second-best is \underline{underlined}.
  Single-domain methods train independently per scenario; federated methods share parameters across domains.
  Due to scenario-specific data filtering (Section~\ref{sec:datasets}), single-domain results may vary across scenarios for the same domain.}
  \label{tab:main_fullrank}
  \small
  \setlength{\tabcolsep}{5pt}
  \begin{tabular}{@{}cl l cc c cc c@{}}
    \toprule
    & & & \multicolumn{2}{c}{\textbf{Single-Domain Discr.}}
      & \textbf{Single-Domain Gen.}
      & \multicolumn{2}{c}{\textbf{Federated Generative}}
      & \\
    \cmidrule(lr){4-5}\cmidrule(lr){6-6}\cmidrule(lr){7-8}
    \textbf{Scenario} & \textbf{Domain} & \textbf{Metric}
      & GRU4Rec & SASRec
      & TIGER
      & TIGER+FedAvg & TIGER+FedProx
      & \ours{} \\
    \midrule
    \multirow{4}{*}{FK}
    & \multirow{2}{*}{Food}
      & H@10 & 0.104 & 0.107 & 0.114 & 0.119 & \second{0.122} & \best{0.179} \\
    & & N@10 & 0.059 & 0.057 & 0.056 & 0.061 & \second{0.069} & \best{0.077} \\
    \cmidrule(l){2-9}
    & \multirow{2}{*}{Kitchen}
      & H@10 & 0.053 & \second{0.055} & 0.033 & 0.033 & 0.034 & \best{0.061} \\
    & & N@10 & \second{0.027} & \second{0.027} & 0.017 & 0.017 & 0.017 & \best{0.029} \\
    \midrule
    \multirow{4}{*}{GS}
    & \multirow{2}{*}{Grocery}
      & H@10 & 0.144 & \second{0.164} & 0.145 & 0.159 & 0.167 & \best{0.175} \\
    & & N@10 & 0.081 & \second{0.092} & 0.089 & 0.095 & 0.095 & \best{0.104} \\
    \cmidrule(l){2-9}
    & \multirow{2}{*}{Sports}
      & H@10 & \second{0.198} & 0.105 & 0.201 & 0.204 & 0.208 & \best{0.216} \\
    & & N@10 & \second{0.110} & 0.081 & 0.079 & 0.081 & 0.089 & \best{0.119} \\
    \midrule
    \multirow{6}{*}{GBS}
    & \multirow{2}{*}{Grocery}
      & H@10 & 0.144 & \best{0.164} & 0.145 & 0.149 & 0.151 & \second{0.157} \\
    & & N@10 & 0.081 & 0.092 & 0.101 & 0.101 & \second{0.112} & \best{0.119} \\
    \cmidrule(l){2-9}
    & \multirow{2}{*}{Beauty}
      & H@10 & \second{0.184} & \best{0.191} & 0.106 & 0.137 & 0.141 & 0.162 \\
    & & N@10 & \best{0.113} & \second{0.111} & 0.055 & 0.076 & 0.062 & 0.098 \\
    \cmidrule(l){2-9}
    & \multirow{2}{*}{Sports}
      & H@10 & \second{0.208} & 0.177 & 0.163 & 0.201 & 0.204 & \best{0.209} \\
    & & N@10 & \second{0.117} & 0.104 & 0.125 & 0.129 & 0.136 & \best{0.146} \\
    \midrule
    \multirow{8}{*}{GKBS}
    & \multirow{2}{*}{Grocery}
      & H@10 & 0.098 & \second{0.111} & 0.101 & 0.115 & 0.118 & \best{0.122} \\
    & & N@10 & 0.054 & 0.059 & 0.055 & 0.059 & \second{0.062} & \best{0.074} \\
    \cmidrule(l){2-9}
    & \multirow{2}{*}{Kitchen}
      & H@10 & 0.039 & \second{0.047} & 0.051 & 0.056 & 0.071 & \best{0.079} \\
    & & N@10 & 0.021 & 0.024 & 0.031 & 0.033 & \second{0.039} & \best{0.045} \\
    \cmidrule(l){2-9}
    & \multirow{2}{*}{Beauty}
      & H@10 & 0.118 & \second{0.136} & 0.122 & 0.135 & \second{0.136} & \best{0.142} \\
    & & N@10 & 0.073 & 0.080 & 0.094 & 0.099 & \second{0.107} & \best{0.122} \\
    \cmidrule(l){2-9}
    & \multirow{2}{*}{Sports}
      & H@10 & 0.120 & 0.124 & 0.129 & 0.134 & \second{0.139} & \best{0.148} \\
    & & N@10 & \second{0.110} & 0.099 & 0.117 & 0.117 & 0.121 & \best{0.125} \\
    \bottomrule
  \end{tabular}
\end{table*}

\begin{table}[t]
  \centering
  \caption{Results under 999-neg sampled evaluation (scenario-level macro-average).
  Best is \textbf{bold}; second-best is \underline{underlined}.}
  \label{tab:main_999neg}
  \resizebox{\columnwidth}{!}{%
  \begin{tabular}{@{}ll ccc cc c@{}}
    \toprule
    & & \multicolumn{3}{c}{\textbf{2-Domain}}
      & \multicolumn{2}{c}{\textbf{3-Domain}}
      & \textbf{4-Dom.} \\
    \cmidrule(lr){3-5}\cmidrule(lr){6-7}\cmidrule(lr){8-8}
    & \textbf{Method}
      & \textbf{FK} & \textbf{GS} & \textbf{GB}
      & \textbf{FKB} & \textbf{GBS}
      & \textbf{GKBS} \\
    \midrule
    \multirow{7}{*}{\rotatebox{90}{H@10}}
    & LocalSASRec    & .130 & .153 & .175 & .151 & .162 & .144 \\
    & LocalCL4SRec   & .134 & .154 & .173 & .155 & .160 & .141 \\
    & LocalDuoRec    & .139 & .155 & .173 & .160 & .168 & .148 \\
    \cmidrule(l){2-8}
    & FedSASRec      & .127 & .149 & .167 & .142 & .148 & .129 \\
    & FedCL4SRec     & .127 & .148 & .171 & .148 & .152 & .132 \\
    & FedDCSR        & \second{.179} & \second{.170} & \second{.194} & \second{.187} & \second{.177} & \second{.180} \\
    \cmidrule(l){2-8}
    & \ours{}        & \best{.187} & \best{.208} & \best{.207} & \best{.200} & \best{.191} & \best{.186} \\
    \midrule
    \multirow{7}{*}{\rotatebox{90}{N@10}}
    & LocalSASRec    & .077 & .096 & .110 & .092 & .102 & .087 \\
    & LocalCL4SRec   & .080 & .095 & .109 & .095 & .100 & .086 \\
    & LocalDuoRec    & .081 & .094 & .109 & .097 & .104 & .089 \\
    \cmidrule(l){2-8}
    & FedSASRec      & .075 & .092 & .107 & .087 & .095 & .079 \\
    & FedCL4SRec     & .074 & .091 & .110 & .089 & .097 & .080 \\
    & FedDCSR        & \second{.098} & \second{.096} & \second{.111} & \second{.104} & \second{.103} & \second{.100} \\
    \cmidrule(l){2-8}
    & \ours{}        & \best{.103} & \best{.112} & \best{.124} & \best{.112} & \best{.111} & \best{.106} \\
    \bottomrule
  \end{tabular}%
  }
\end{table}


\subsubsection{Full-ranking analysis (Table~\ref{tab:main_fullrank}).}

\textbf{\ours{} is the strongest federated generative method.}
Across all 22 domain-metric cells, \ours{} consistently outperforms both TIGER+FedAvg and TIGER+ FedProx.
The improvement is most pronounced on GKBS, the most heterogeneous scenario:
on GKBS Grocery N@10, \ours{} achieves 0.074, improving over TIGER+FedProx (0.062, +19.4\%) and TIGER+FedAvg (0.059, +25.4\%);
on GKBS Kitchen H@10, \ours{} reaches 0.079 versus TIGER+FedProx (0.071) and TIGER+FedAvg (0.056).
These gains confirm that prototype-personalized aggregation and reliability-aware CF residuals provide consistent benefits over na\"ive federation strategies.

\textbf{Federation progressively improves generative models.}
A clear progression from TIGER $\rightarrow$ TIGER+FedAvg $\rightarrow$ TIGER+FedProx $\rightarrow$ \ours{} is visible in most domains.
On GKBS Kitchen H@10: 0.051 $\rightarrow$ 0.056 $\rightarrow$ 0.071 $\rightarrow$ 0.079;
on GS Grocery H@10: 0.145 $\rightarrow$ 0.159 $\rightarrow$ 0.167 $\rightarrow$ 0.175;
on FK Kitchen H@10: 0.033 $\rightarrow$ 0.033 $\rightarrow$ 0.034 $\rightarrow$ 0.061.
This demonstrates that (1)~cross-domain knowledge transfer benefits SID-based generation,
(2)~proximal regularization helps but is insufficient under domain heterogeneity,
and (3)~the reliability-aware interface and personalized aggregation of \ours{} provide further substantial gains.

\textbf{\ours{} is competitive with non-federated discriminative baselines.}
GRU4Rec and SASRec train independently with full access to local interactions.
\ours{} achieves the best result in the majority of domain-metric cells, including all eight cells in GKBS, all four in FK, and both domains in GS.
On FK Kitchen, \ours{} (H@10: 0.061) surpasses both SASRec (0.055) and GRU4Rec (0.053), despite TIGER and its federated variants falling well below discriminative methods on this domain.
In domains with dense interactions and strong collaborative signals, such as GBS Beauty (SASRec H@10: 0.191 vs.\ \ours{} 0.162), discriminative methods retain an advantage, as embedding-based scoring benefits directly from rich interaction data.
Crucially, \ours{} achieves this competitive performance while keeping all raw user interactions private---a constraint that non-federated methods do not face.

\subsubsection{999-neg analysis (Table~\ref{tab:main_999neg}).}

Under the sampled evaluation protocol adopted by prior federated CDR methods, \ours{} achieves the best results across all six scenarios on both metrics.

\textbf{\ours{} consistently outperforms all federated baselines.}
Compared with FedDCSR, the strongest baseline, \ours{} improves H@10 from 0.180 to 0.186 (+3.3\%) on GKBS and from 0.170 to 0.208 (+22.4\%) on GS.
The improvement is largest on GS, where the low behavioral affinity between Grocery and Sports makes uniform aggregation particularly harmful.

\textbf{Na\"ive federation causes negative transfer.}
FedSASRec and FedCL4SRec consistently underperform their non-federated counterparts: on GKBS, H@10 drops from 0.144 (LocalSASRec) to 0.129 (FedSASRec) after federation.
This confirms that negative transfer is a general challenge in federated CDR, not specific to generative models.
FedDCSR partially addresses this through disentangled representations, but \ours{} achieves further gains by combining a unified SID vocabulary with prototype-personalized aggregation, yielding consistent improvements across scenarios of varying size and domain affinity.

\subsection{Ablation Study (RQ2)}
\label{sec:ablation}

We ablate the two design principles of \ours{}: the reliability-aware semantic interface and the prototype-personalized generator.
We select FK (high-affinity, 2-domain) and GKBS (high-heterogeneity, 4-domain) to examine component contributions under contrasting conditions.
Table~\ref{tab:ablation} reports N@10 (scenario-level macro-average).

\begin{table}[t]
  \caption{Ablation study (N@10, scenario-level macro-average).
  $\Delta$ is the relative change versus the full model, averaged over both scenarios.}
  \label{tab:ablation}
  \centering
  \small
  \begin{tabular*}{\columnwidth}{@{\extracolsep{\fill}}lccc@{}}
    \toprule
    \textbf{Variant} & \textbf{FK} & \textbf{GKBS} & \textbf{$\Delta$ avg} \\
    \midrule
    \ours{} full & \best{0.048} & \best{0.092} & -- \\
    \midrule
    \multicolumn{4}{@{}l}{\emph{Reliability-aware semantic interface}} \\
    \midrule
    w/o CF residual (SID only)        & 0.042 & 0.080 & $-$12.8\% \\
    w/o item confidence $\rho_v$      & 0.044 & 0.085 & $-$8.0\% \\
    w/o domain gate $g_i$             & 0.045 & 0.086 & $-$6.4\% \\
    w/o dense head $f_{\mathrm{head},i}$ & 0.046 & 0.089 & $-$3.7\% \\
    w/o confidence-weighted dense loss & 0.046 & 0.087 & $-$4.8\% \\
    \midrule
    \multicolumn{4}{@{}l}{\emph{Prototype-personalized generator}} \\
    \midrule
    FedAvg aggregation$^\dagger$      & 0.044 & 0.079 & $-$11.2\% \\
    w/o prototype weights$^\ddagger$  & 0.044 & 0.081 & $-$10.1\% \\
    w/o shared-private split (shared FFN only) & 0.045 & 0.084 & $-$7.5\% \\
    w/o private expert                & 0.046 & 0.088 & $-$4.3\% \\
    \midrule
    \multicolumn{4}{@{}l}{\emph{Combined}} \\
    \midrule
    w/o CF residual + FedAvg$^\dagger$ & 0.041 & 0.078 & $-$14.9\% \\
    \bottomrule
  \end{tabular*}
  \vspace{-4mm}
\end{table}

\textbf{Reliability-aware semantic interface.}
Removing the CF residual entirely (SID only) causes the largest drop in this group ($-$12.8\% on average), confirming that local collaborative signals provide complementary information beyond item-side semantics.
Removing item confidence $\rho_v$ or the domain gate $g_i$ weakens the model's ability to distinguish reliable from noisy CF signals: without $\rho_v$, long-tail items inject unreliable embeddings; without $g_i$, the model cannot calibrate the overall CF contribution at the domain level.
The dense head and confidence-weighted loss each contribute moderately ($-$3.7\% and $-$4.8\%), validating the auxiliary alignment mechanism that encourages the shared encoder to preserve reliable CF evidence during training.

\textbf{Prototype-personalized generator.}
We distinguish two aggregation ablations.
FedAvg aggregation ($\dagger$) replaces both personalized weighting and the shared-private split with a single data-volume-weighted global model, averaging all experts across clients.
In contrast, w/o prototype weights ($\ddagger$) keeps the shared-private MoE structure with private experts remaining local, but sets similarity weights uniform ($\tau_a\!\to\!\infty$), so only shared parameters are averaged by data volume.
Replacing the entire personalized pipeline with standard FedAvg causes a substantial drop ($-$11.2\%).
The degradation is more pronounced on GKBS (0.079 vs.\ 0.092, $-$14.1\%) than on FK (0.044 vs.\ 0.048, $-$8.3\%), consistent with the expectation that personalized aggregation matters more when domains are behaviorally diverse.
Keeping the shared-private MoE structure but removing only the cosine similarity weights (w/o prototype weights) recovers part of the loss (GKBS: 0.081 vs.\ 0.079), confirming that similarity-aware aggregation and domain-specific capacity are complementary rather than redundant.

\textbf{Division of labor between the two principles.}
The ablations reveal a cross-over pattern that clarifies the respective roles of the two design principles.
On the high-affinity FK scenario, removing local CF evidence ($-$12.5\%) is more harmful than replacing personalized aggregation with FedAvg ($-$8.3\%), indicating that once domains already share a useful semantic language, the main bottleneck becomes recovering domain-specific collaborative behavior.
On the heterogeneous GKBS scenario, the pattern reverses: FedAvg ($-$14.1\%) is more harmful than CF removal ($-$13.0\%), showing that negative transfer becomes the binding constraint as domain relatedness decreases.
This cross-over supports the central thesis of \ours{}: a stable item language needs both local evidence interfaces and personalized sharing, and their relative importance is governed by the domain-relatedness regime.

\textbf{Combined ablation.}
Removing both the CF residual and personalized aggregation simultaneously yields the lowest performance ($-$14.9\%), approaching TIGER+FedAvg in Table~\ref{tab:main_fullrank}.
This confirms that the two principles are complementary: the semantic interface recovers local behavioral evidence, while personalized federation reduces negative transfer, and their combination produces gains beyond what either achieves alone.

\subsection{Cold-Start Analysis (RQ3)}
\label{sec:coldstart}
\begin{figure}[t]
  \centering
  \includegraphics[width=\columnwidth]{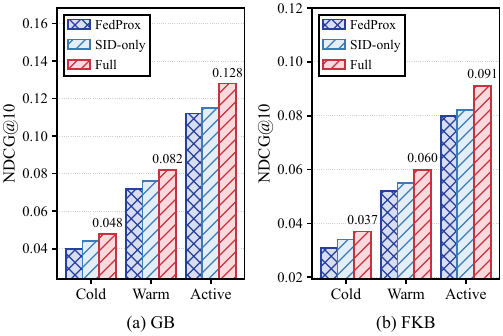}
  \caption{Cold-start analysis: NDCG@10 across user activity groups on (a)~GB and (b)~FKB. The stable SID vocabulary provides the main cold-start gain; the reliability-aware CF residual adds an increasing benefit as user activity grows.}
  \label{fig:coldstart}
  \vspace{-4mm}
\end{figure}

Since cold-start alleviation is a central motivation for CDR, we examine how \ours{} performs across users with different activity levels.
We partition test users into three groups by training sequence length: \emph{cold} ($\le 7$), \emph{warm} ($8$--$15$), and \emph{active} ($\ge 16$).
On GB, the cold/warm/active groups account for 45\%, 28\%, and 27\% of test users, respectively; on FKB, the proportions are 57\%, 25\%, and 18\%.
Figure~\ref{fig:coldstart} reports NDCG@10 on GB and FKB for \ours{}, its SID-only ablation (w/o CF residual), and TIGER+FedProx.

The results separate two sources of benefit.
The first source is the stable SID vocabulary.
On cold users, both \ours{} variants outperform TIGER+FedProx, confirming that the semantic item language provides a strong cold-start foundation because item representations are grounded in metadata rather than interaction counts.
The SID-only variant does not degrade on cold users relative to TIGER+FedProx, showing that the reliability modulation ($\rho_v$ and $g_i$) successfully suppresses noisy CF signals rather than injecting harmful noise.
The second source is the reliability-aware CF residual, whose benefit grows with user activity: the gap between \ours{} full and the SID-only variant widens from cold to active users because richer interaction histories make local CF embeddings more reliable.
Thus, \ours{} does not rely on CF signals for cold users; instead, it falls back to the semantic item language and progressively activates CF evidence as it becomes trustworthy.

\subsection{Sensitivity and Diagnostic Analysis (RQ4)}
\label{sec:analysis}
We further analyze how \ours{} behaves under different reliability and personalization settings.

\subsubsection{Hyper-parameter sensitivity.}
\begin{figure}[t]
  \centering
  \includegraphics[width=\columnwidth]{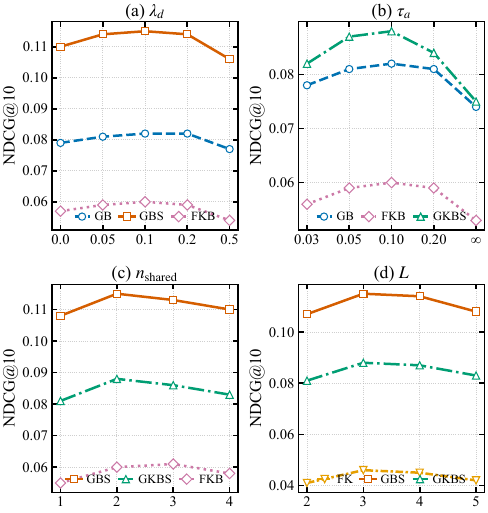}
  \caption{Sensitivity analysis: (a)~NDCG@10 vs.\ dense loss weight $\lambda_d$, (b)~NDCG@10 vs.\ aggregation temperature $\tau_a$, (c)~NDCG@10 vs.\ number of shared experts $n_{\text{shared}}$, and (d)~NDCG@10 vs.\ number of RQ levels $L$. The dense loss weight and aggregation temperature exhibit a clear optimum around 0.10, while performance is relatively stable across $n_{\text{shared}}\in\{2,3,4\}$ and $L\in\{3,4\}$.}
  \label{fig:sensitivity}
\end{figure}

Figure~\ref{fig:sensitivity} reports validation NDCG@10 under varying dense loss weights $\lambda_d$, aggregation temperatures $\tau_a$, numbers of shared experts $n_{\text{shared}}$, and RQ levels $L$.

The dense loss weight $\lambda_d$ shows a clear optimum around 0.10; too small a weight loses the alignment benefit, while too large a weight causes the auxiliary objective to dominate the SID generation loss.
The aggregation temperature $\tau_a$ also exhibits a moderate optimum at 0.10: very low temperatures concentrate aggregation weights on the single most similar domain and reduce diversity, while $\tau_a=\infty$ collapses to data-volume averaging and suffers from negative transfer.
Notably, the performance gap between $\tau_a=0.10$ and $\tau_a=\infty$ is larger on the four-domain scenario GKBS (14.8\% relative drop) than on GB (11.0\%), confirming that personalized aggregation becomes more important as domain heterogeneity increases.
The number of shared experts $n_{\text{shared}}$ and the number of RQ levels $L$ are relatively robust: performance is stable across $n_{\text{shared}}\in\{2,3,4\}$ and $L\in\{3,4\}$, with slight degradation at extreme values.

\subsubsection{Diagnostic statistics.}
Table~\ref{tab:gate_diagnostic} summarizes the learned gates and confidence statistics after convergence.

\begin{table}[t]
  \caption{Diagnostic statistics for reliability-aware fusion. $g_i$ is the learned client-local gate; $\bar{\rho}$ is the mean item confidence.}
  \label{tab:gate_diagnostic}
  \centering
  \small
  \begin{tabular*}{\columnwidth}{@{\extracolsep{\fill}}lccc@{}}
    \toprule
    \textbf{Domain} & $g_i$ & $\bar{\rho}$ & \textbf{Interpretation} \\
    \midrule
    Food    & 0.67 & 0.42 & Strong CF quality \\
    Grocery & 0.61 & 0.38 & Moderate-strong CF quality \\
    Kitchen & 0.58 & 0.35 & Moderate CF quality \\
    Beauty  & 0.52 & 0.31 & Moderate CF quality \\
    Sports  & 0.41 & 0.25 & Sparse / noisy CF \\
    \bottomrule
  \end{tabular*}
\end{table}

Domains with higher average interactions per item---Food (26.4), Grocery (21.9)---learn higher gate values and item confidence, allowing more CF evidence into the SID representation.
Sparser domains such as Sports (13.3 avg interactions per item) learn lower gates, relying more on the SID-only representation.
This confirms that the gate $g_i$ and confidence $\rho_v$ learn to calibrate the domain-level CF contribution without manual tuning, consistent with the design motivation of the reliability-aware interface.

\subsection{Efficiency Analysis (RQ5)}
\label{sec:efficiency}
\begin{figure}[t]
  \centering
  \includegraphics[width=\columnwidth]{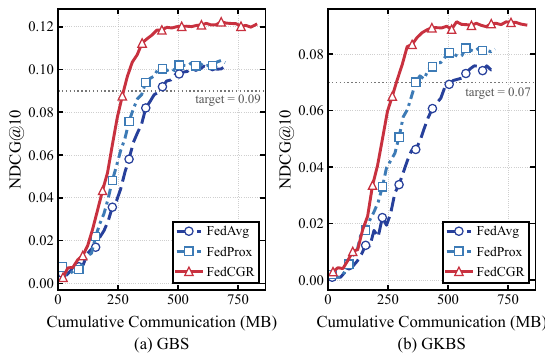}
  \caption{NDCG@10 vs.\ cumulative upload communication on (a)~GBS and (b)~GKBS. Dashed lines mark reference targets. Despite higher per-round cost, \ours{} reaches each target with less total communication than both baselines.}
  \label{fig:comm_efficiency}
  \vspace{-8mm}
\end{figure}

Figure~\ref{fig:comm_efficiency} plots NDCG@10 against cumulative upload communication on GBS and GKBS.

\textbf{Per-round cost.}
\ours{} uploads 5.1M shared parameters (20.6~MB) per client per round, approximately 18\% more than TIGER+FedAvg and TIGER+FedProx (4.3M, 17.4~MB), due to shared experts and the dense auxiliary loss.
The prototype upload is negligible ($<$1~KB).
Download communication equals the upload size for all three methods: TIGER+FedAvg and TIGER+FedProx broadcast one global model (4.3M), while \ours{} sends each client a personalized aggregate of the same shared parameter size (5.1M).
Per-round training time is 4.3~min for \ours{} vs.\ 3.5~min for TIGER+FedAvg.
The reported time excludes one-time local CF pre-training (approximately 8~minutes per domain), which is performed independently before federated generator training.

\textbf{Accuracy-per-communication.}
The relevant efficiency metric is not per-round payload alone, but how much total communication is needed to reach a given accuracy target.
Although \ours{} costs more per round, it converges faster and reaches a higher plateau.
As shown in Figure~\ref{fig:comm_efficiency}(a), on GBS \ours{} reaches the 0.09 N@10 target at round~14 with 288~MB cumulative upload, whereas TIGER+FedAvg requires 435~MB (round~25) and TIGER+FedProx requires 365~MB (round~21)---a 34\% reduction relative to FedAvg.
On the more heterogeneous GKBS (Figure~\ref{fig:comm_efficiency}(b)), \ours{} crosses the 0.07 target at 288~MB (round~14), while TIGER+FedAvg needs 522~MB (round~30) and TIGER+FedProx needs 365~MB (round~21)---a 45\% reduction relative to FedAvg.
The communication savings grow with domain heterogeneity, confirming that prototype-personalized aggregation not only improves accuracy but also accelerates convergence, turning a modest per-round overhead into a net reduction in total communication budget.

\section{Related Work}

\textbf{CDR and Federated CDR.} CDR transfers knowledge across domains to alleviate sparsity and cold-start problems, from early shared-factor or mapping-based methods~\cite{cmf,emcdr} to neural transfer models such as CoNet~\cite{conet}, DDTCDR~\cite{ddtcdr}, and RecGURU~\cite{recguru}, as well as graph-based~\cite{bitgcf,ppgn} and contrastive~\cite{ccdr} cross-domain methods. Recent studies further address distribution shift and negative transfer through source filtering, causal disentanglement, and robust optimization~\cite{cut,hjid,uiea}, while GenCDR~\cite{gencdr} introduces unified SIDs for centralized generative CDR. Federated CDR instead keeps raw interactions local~\cite{fedavg}, with existing methods exploring decentralized transfer and privacy protection~\cite{fedct,fedcdr}, disentangled or graph-based representation learning~\cite{feddcsr,fedhcdr,fedgcdr_neurips,p2fcdr,fedpclcdr}, and semantic bridging without overlapping users~\cite{pfcr,ffmsr,fedcrf,fedcsr,fedecider}. General personalized FL has also studied regularization-based, shared-representation, and clustered aggregation strategies~\cite{pfedme,ditto,fedrep,fedpac,ifca}. Unlike these mostly discriminative approaches, \ours{} studies federated CDR as autoregressive generation over shared SIDs, where item alignment is induced by a stable semantic vocabulary before federated training.

\textbf{Generative Recommendation with Semantic IDs.} Generative recommendation formulates retrieval as sequence generation over discrete item tokens~\cite{survey-sid,survey-sid1,genrec}, and has been extended with multimodal item representations~\cite{mmsr}. VQ-Rec~\cite{vqrec} learns vector-quantized transferable item codes, and TIGER~\cite{tiger} establishes RQ-VAE-based SID generation for next-item recommendation. Later work improves SID construction through end-to-end, contrastive, hierarchical, textual, or industrial-scale alignment objectives~\cite{lmindexer,lcrec,letter,idgenrec,das}, and extends the paradigm with order-agnostic identifiers, parallel SID prediction, temporal modeling, speculative decoding, and dense-generative retrieval unification~\cite{setrec,rpg,grut,atspeed,liger}. GenCDR~\cite{gencdr} applies SID-based generation to centralized CDR. In contrast, \ours{} targets federated deployment, where the tokenizer must remain fixed, CF evidence is local, and aggregation must handle domain heterogeneity; we address these constraints with reliability-aware CF injection and prototype-personalized federated generation.

\section{Conclusion}
This paper proposed \ours{}, which formulates federated CDR as generation over a stable semantic item language.
A reliability-aware semantic interface injects local CF evidence into fixed SID representations through confidence-gated residuals; a prototype-personalized generator selectively aggregates shared parameters according to domain relatedness.
Experiments on six Amazon scenarios confirm consistent improvements over federated generative baselines and competitive performance against strong sequential and federated CDR methods.
Ablations reveal a cross-over pattern: CF enrichment dominates in high-affinity scenarios while personalized aggregation becomes the binding constraint under high heterogeneity.
Despite an 18\% higher per-round communication, faster convergence reduces the total budget by 34--45\%.
Future work may integrate secure aggregation, learn confidence estimates end-to-end, and extend the framework to scenarios where new domains or items arrive incrementally during federation.

\clearpage

\section*{Generative AI Disclosure}
Generative AI tools were used only for language polishing and did not contribute to the research ideas, method design, experiments, analysis, or conclusions.

\bibliographystyle{ACM-Reference-Format}
\bibliography{references}

\end{document}